\documentclass[10pt,journal,compsoc]{IEEEtran}

\ifCLASSOPTIONcompsoc
  \usepackage[nocompress]{cite}
\else
  \usepackage{cite}
\fi

\ifCLASSINFOpdf
\else
\fi

\usepackage{graphicx}
\usepackage[cmex10]{amsmath}
\usepackage{subfig}
\usepackage{array}
\usepackage{booktabs}
\usepackage{multirow}
\usepackage{threeparttable}
\usepackage{amsfonts}
\usepackage{tabularx}

\newcolumntype{Y}{>{\centering\arraybackslash}X}
\newcolumntype{b}{>{\centering\arraybackslash}X}
\newcolumntype{s}{>{\hsize=.33\hsize\centering\arraybackslash}X}

\makeatletter
\newcommand*{\rom}[1]{\expandafter\@slowromancap\romannumeral #1@}
\makeatother

\begin{document}
\title{Visual Semantic Information Pursuit: A Survey}

\author{Daqi~Liu,~Miroslaw~Bober,~\IEEEmembership{Member,~IEEE},~Josef~Kittler,~\IEEEmembership{Life Member,~IEEE}% <-this % stops a space
\IEEEcompsocitemizethanks{\IEEEcompsocthanksitem The authors are with the Centre for Vision, Speech and Signal Processing, University of Surrey, Guildford GU2 7XH, U.K. \protect\\
E-mail: \{daqi.liu, m.bober, j.kittler\}@surrey.ac.uk 
\protect\\ 
\protect\\ Preliminary work. Under review by IEEE Transactions on Pattern Analysis and Machine Intelligence (PAMI). Do not distribute.} 
}

%{Shell \MakeLowercase{\textit{et al.}}: Bare Demo of IEEEtran.cls for Computer Society Journals}

\IEEEtitleabstractindextext{%
\begin{abstract}
Visual semantic information comprises two important parts: the meaning of each visual semantic unit and the coherent visual semantic relation conveyed by these visual semantic units. Essentially, the former one is a visual perception task while the latter one corresponds to visual context reasoning. Remarkable advances in visual perception have been achieved due to the success of deep learning. In contrast, visual semantic information pursuit, a visual scene semantic interpretation task combining visual perception and visual context reasoning, is still in its early stage. It is the core task of many different computer vision applications, such as object detection, visual semantic segmentation, visual relationship detection or scene graph generation. Since it helps to enhance the accuracy and the consistency of the resulting interpretation, visual context reasoning is often incorporated with visual perception in current deep end-to-end visual semantic information pursuit methods. However, a comprehensive review for this exciting area is still lacking. In this survey, we present a unified theoretical paradigm for all these methods, followed by an overview of the major developments and the future trends in each potential direction. The common benchmark datasets, the evaluation metrics and the comparisons of the corresponding methods are also introduced.
\end{abstract}

}

\maketitle

\IEEEdisplaynontitleabstractindextext

% \fi
\IEEEpeerreviewmaketitle

\IEEEraisesectionheading{\section{Introduction}\label{sec:introduction}}

\IEEEPARstart{S}{emantics} is the linguistic and philosophical study of meaning, in language, programming languages or formal logics. In linguistics, the semantic signifiers can be words, phrases, sentences or paragraphs. To interpret the complicated signifiers  such as phrases or sentences, we need to understand the meaning of each word as well as the semantic relation among those words. Here, words are the basic semantic units and the semantic relation is any relationship between two or more words based on the meaning of the words. In other words, the semantic relation defines the consistency among the associated semantic units in terms of meaning, which guarantee that the corresponding complex semantic signifier can be interpreted.


The above strategy can be seamlessly applied to the visual semantic information pursuit, in which the basic semantic units are potential pixels or potential bounding boxes while the visual semantic relation is represented as the local visual relationship structure or the holistic scene graph. For visual perception tasks such as visual semantic segmentation or object detection, the visual semantic relation promotes smoothness and consistency among the input visual semantic units. It acts as a regularizer and causes the associated visual semantic units to be biased towards certain configurations which are more likely to occur. For visual context reasoning applications, such as visual relationship detection or scene graph generation, the corresponding visual semantic units are considered as the associated context information and different inference methods are applied to pursue the visual semantic relation. In a word, the visual semantic units and the visual semantic relation are complementary. The visual semantic units are the prerequisites of the visual semantic relation, while the visual semantic relation can be explored to further improve the detection accuracy of the visual semantic units. 

The extent to which a visual semantic information pursuit method can interpret the input visual stimuli is totally dependent on the prior knowledge of the observer. Vocabulary is one part of the knowledge, which defines the meaning of each visual semantic unit. The vocabulary itself may be enough for some specific visual perception tasks (such as weakly-supervised learning for object detection). However, for visual context reasoning applications, it is certainly not sufficient since we still need additional knowledge to identify and understand the interpretable visual semantic relations. In most cases, besides the vocabulary, the associated benchmark datasets should provide other ground-truth information about the visual semantic relations.

In this survey, four main research topics in visual semantic information pursuit are introduced: object detection, visual semantic segmentation, visual relationship detection and scene graph generation. Traditionally, the first two tasks are categorized as visual perception applications. Instead of considering them as a single visual perception task, the current visual semantic information pursuit research treats them as a combination of perception and reasoning. Therefore, unlike the previous surveys, all applications mentioned in this article include visual context reasoning modules and can be trained end-to-end through the associated deep learning models.

Specifically, object detection (OD) aims at detecting all possible objects appearing in the input image by assigning corresponding bounding boxes as well as their associated labels. Essentially, it consists of two modules: localization and classification. The former is achieved by an associated regression algorithm while the latter is typically implemented by a corresponding classification method. In this article, we only focus on introducing the object detection methods with visual context reasoning modules \cite{chen2017spatial}, \cite{chen2018iterative}, \cite{zeng2017crafting}, \cite{liu2018structure}. The comprehensive survey of conventional object detection methods can be found in paper \cite{han2018advanced}.

Visual semantic segmentation (VSS) \cite{zheng2015conditional}, \cite{liu2017deep}, \cite{shuai2018scene} refers to labelling each pixel to be one of the semantic categories. To robustly parse input images, effective visual context modelling is essential. Due to its intrinsic characteristic, visual semantic segmentation is often formulated as an undirected graphical model, such as Undirected Cyclic Graph (UCG) \cite{shuai2018scene} or Conditional Random Field (CRF) \cite{lafferty2001conditional}. In most cases, the energy function corresponding to the undirected graphical model is factorized into two potential functions: unary function and binary function. The former generates the predicted label for each input pixel while the latter defines the pairwise interaction between adjacent pixels. As a constraint term, the binary potential function is used to regulate the predicted labels generated from the unary potential function to be spatially consistent and smooth.
\begin{figure}[!t]
\centering
\includegraphics[width=\linewidth]{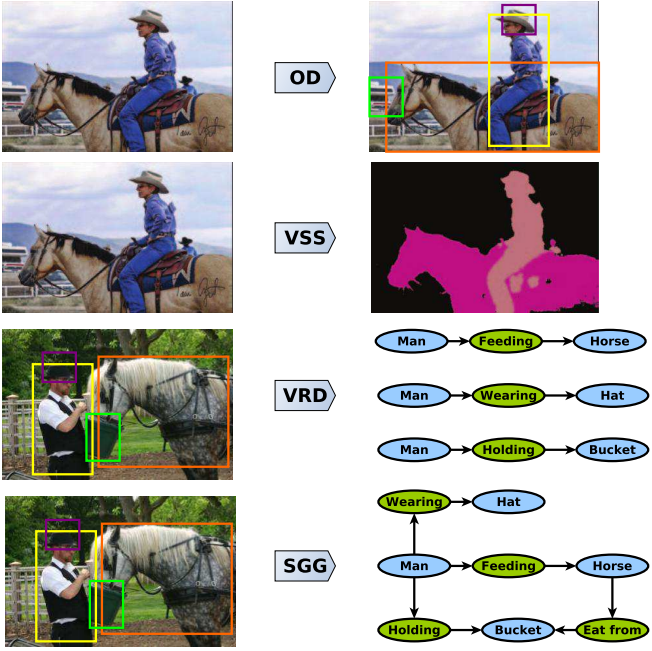}
\caption{Four main visual semantic information pursuit applications are introduced in this survey, which include object detection (OD), visual semantic segmentation (VSS), visual relationship detection (VRD) and scene graph generation (SGG).}
\end{figure} 

Visual relationship detection (VRD) \cite{lu2016visual}, \cite{dai2017detecting}, \cite{li2017vip}, \cite{liang2017deep}, \cite{yu2017visual}, \cite{zhang2017visual}, \cite{zhang2017relationship}, \cite{plesse2018visual}, \cite{zhang2018large} focuses on recognizing the potential relationship between pairs of detected objects, in which the output is often formulated as a triplet in the form of $(subject, predicate, object)$. Generally, it is not sufficient to interpret the input image by only recognizing the individual objects. The visual relationship triplet, in particular the predicate, plays an important role in understanding the input images. However, it is often hard to predict the predicates since they tend to exhibit a long-tail distribution. In most cases, for a same predicate, the diversity of the subject-object combinations is often enormous \cite{zhuang2017towards}. In other words, compared with the individual objects, the corresponding predicates capture more general abstractions from the input images.

Scene graph generation (SGG) \cite{xu2017scene}, \cite{li2017scene}, \cite{zellers2018neural}, \cite{yang2018graph}, \cite{li2018factorizable}, \cite{herzig2018mapping}, \cite{woo2018linknet} builds a visually-grounded scene graph to explicitly model the objects and their relationships. Unlike the visual relationship detection, the scene graph generation aims to build a global scene graph instead of producing local visual relationship triplets. The contextual information conveyed within the scene graph is not limited to isolated triplets, but extends to all the related objects and predicates. To jointly infer the scene graph, message passing among the associated objects and predicates is essential in scene graph generation tasks.
 
The above visual semantic information pursuit applications, as shown in Fig.1, try to interpret the input image at different semantic levels. For instance, object detection tries to interpret the visual semantic units while visual semantic segmentation seeks to interpret the visual semantic regions (essentially, semantic regions are semantic units with different representation forms); Visual relationship detection tries to interpret the visual semantic phrases while the scene graph generation attempts to interpret the visual semantic scene. Visual semantic information from the above low- and mid-level visual intelligence tasks is the basis of the high-level visual intelligence tasks such as visual captioning \cite{karpathy2015deep}, \cite{he2017deep}, \cite{fu2017aligning} or visual question answering \cite{malinowski2017ask}, \cite{agrawal2017vqa}, \cite{teney2017visual}.

Specifically, to accomplish the visual semantic information pursuit, three key questions need to be answered: 1) What kind of visual context information is required? 2) How to model the required visual context information? 3) How to infer the posterior distribution given the visual context information? This article presents a comprehensive survey of the state-of-the-art visual semantic information pursuit algorithms, which try to answer the above questions. 

This survey is organized as follows: Section \rom{2} presents the terminologies and the fundamentals of the visual semantic information pursuit. Section \rom{3} introduces a unified paradigm for all visual semantic information pursuit methods. The major developments and the future research directions are covered in Section \rom{4} and Section \rom{5}, respectively. The common benchmarks, the evaluation metrics and the experimental comparison of the key methods are summarized in Section \rom{6} and Section \rom{7}, respectively. Finally, the conclusions are drawn in Section \rom{8}.

\section{preliminary knowledge}

A typical visual semantic information pursuit method consists of two modules: a visual perception module and a visual context reasoning module. The visual perception module tries to detect visual semantic units from the input visual stimuli and assign specific meaning to them. Convolutional neural network (CNN) architectures such as fully convolutional networks (FCNs) \cite{long2015fully} or faster regional CNNs (faster R-CNNs) \cite{ren2015faster} are often used to model the visual perception tasks. They can not only provide the initial predictions of the visual semantic units, but also the locations of the associated bounding boxes. The comprehensive introduction to these CNN models can be found in the previous surveys \cite{han2018advanced}, \cite{arnab2018conditional}. In general, regional proposal networks (RPNs) \cite{ren2015faster}, \cite{girshick2015fast} are often used to produce the proposal bounding boxes, and region of interest (ROI) pooling \cite{ren2015faster} or bilinear feature interpolation \cite{zhang2017visual} are usually applied to obtain the corresponding feature vectors. Three possible prior factors - visual appearance, class information and relative spatial relationship - are often considered in forming the visual semantic perception module, as shown in Fig.2.
\begin{figure}[!t]
\centering
\includegraphics[width=\linewidth]{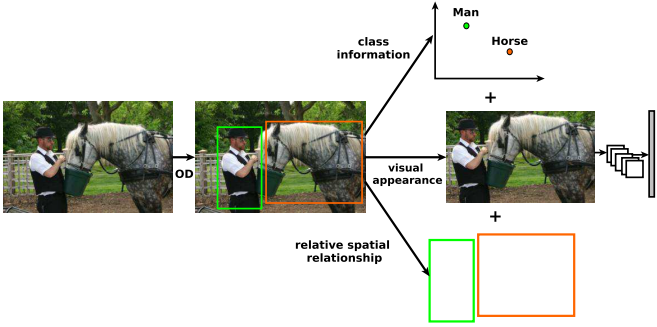}
\caption{Three possible prior factors are often considered in the current visual perception modules.}
\end{figure} 

Given the detected visual semantic units, the aim of the visual context reasoning module is to produce the most probable interpretation through a maximum a posteriori (MAP) inference. Basically, the above MAP inference is a NP-hard integer programming problem \cite{werner2007linear}. However, it is possible to address the above integer programming problem via a linear relaxation, where the generated linear programming problem can be presented as a variational free energy minimization approximation. Within current visual semantic information pursuit methods, to accomplish the MAP inference, the associated marginal polytopes corresponding to the target variational free energies are often approximated by their corresponding feasible polytopes \cite{liu2013variational}. Such feasible polytopes can be further factorized into numerous regions accordingly, in which they can be trained sequentially or in parallel using the corresponding optimization methods. Essentially, the aim of the above approximation is to find a upper bound for the target variational free energy. The tighter the upper bound, the better the MAP inference will be. 

Furthermore, to accomplish the MAP inference, the visual context reasoning module often incorporates prior knowledge to regularize the target variational free energy. Generally, there are two types of prior knowledge: internal prior knowledge and external prior knowledge. The internal prior knowledge is acquired from the visual stimulus itself. For instance, the adjacency of the visual stimuli is a typical internal prior knowledge, i.e.  the adjacent objects tend to have a relationship or the adjacent pixels tend to have the same label. The external prior knowledge is obtained from the external sources, such as tasks, contexts, knowledge bases. For instance, the linguistic knowledge bases like word2vec \cite{mikolov2013efficient}, \cite{mikolov2013distributed} are often used as external prior knowledge since the embeddings can be used to measure the semantic similarity among different words. Specifically, the word embeddings are positioned in the vector space such that words that share common contexts in the corpus are located in close proximity to one another. Accordingly, the current visual semantic pursuit algorithms can generally be divided into the following two categories: bottom-up methods and top-down methods. The former only use internal prior knowledge while the latter incorporate both internal and external prior knowledge. 

Besides the above MAP inference step, one still needs model selection step to resolve the visual semantic information pursuit applications. Specifically, MAP inference step finds the most probable interpretation for the input visual stimuli while model selection step aims to find the best model (an optimum member within a distribution family) through maximizing the corresponding conditional likelihood. Within the current visual semantic information pursuit methods, deep learning-based message passing optimization strategies are often used to accomplish the MAP inference step, while the model selection step is generally implemented by stochastic gradient descent (SGD) methods.  

\section{unified paradigm} 

Within a visual semantic information pursuit system, the visual perception module initializes the visual context reasoning module, while the visual context reasoning module constraints the visual perception module. Those two modules are complementary since both can provide contextual information to each other. In recent years, the deep learning models like the CNNs have been shown to achieve superior performance in numerous visual perception tasks \cite{krizhevsky2012imagenet}, \cite{russakovsky2015imagenet}, \cite{he2016deep}. They become the de facto choices as visual perception modules in the current research. However, the conventional CNNs are still not close to solving the inference tasks within the visual context reasoning modules.

\subsection{Formulation} 

To accomplish an inference task, a probabilistic graphical model is often adopted as a visual semantic information pursuit framework. It uses a graph-based representation as the foundation for encoding a distribution over a multi-dimensional space. It is a factorized representation of the set of independences that hold in a specific distribution. Two types of graphical models are commonly used, namely, Bayesian Networks and Markov Random Fields. The former are directed acyclic graphical models with causality connections while the latter are undirected graphical models with cycles in most cases. Both of them can be reformulated as the corresponding factor graph models. Specifically, within the associated factor graphical model, the visual semantic units are represented as the variable nodes while the visual semantic relations are depicted as the factor nodes. 

Given the associated factor graphical model, the aim of the visual context reasoning module is to infer the most probable interpretation given the observed input visual stimuli. In other words, given the input images and other ground-truth information (such as the locations of the associated bounding boxes), we want to maximize the corresponding posterior distribution. The above MAP inference is a NP-hard integer programming problem and it is often reformulated as a linear programming problem via a linear relaxation \cite{liu2013variational}. Furthermore, within the probabilistic graphical model, the posterior can be derived from the corresponding energy functions according to Boltzmann's Law. Basically, the lower the energy, the more probable the potential interpretation will be. The above linear programming problem can be further expressed as a variational free energy minimization problem. In a word, instead of exact inference, the visual context reasoning module would interpret the input visual stimuli using a relevant variational free energy minimization approximation.

\subsection{Unified Paradigm}

Based on the above analysis, the current visual semantic information pursuit methods follow a unified paradigm. Specifically, the visual perception module applies a corresponding CNN model to produce the visual semantic units, while the visual context reasoning module uses a relevant deep learning-based variational free energy minimization method to approximate the target visual semantic relations from the above visual semantic units. Fig.3 schematically represents the unified paradigm of the current visual semantic information pursuit methods.
\begin{figure}[!t]
\centering
\includegraphics[width=\linewidth]{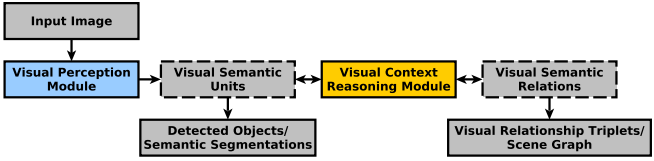}
\caption{The unified paradigm of the current visual semantic information pursuit methods.}
\end{figure} 
 
In this survey, we use the more general scene graph generation task as an example to develop a mathematical model corresponding to the unified paradigm. Other visual semantic information pursuit applications are special cases of this formulation. To generate a visually-grounded scene graph, the corresponding visual perception module applies a CNN model like RPN to automatically obtain an initial set of object bounding boxes $B_{I}$ from the input image $I$. For each proposal bounding box, the visual context reasoning module needs to infer three variables: 1) the associated object class label; 2) the corresponding four bounding box offsets relative to the proposal box coordinates; 3) the relevant predicate labels between the potential object pairs.

Given a set of object classes $\mathcal{C}$ and a set of relationship types $\mathcal{R}$, the above set of variables can be depicted as $X=\{x_{i}^{cls}, x_{i}^{bbox}, x_{i\rightarrow j}|i=1\cdots n, j=1\cdots n, i\neq j \}$, where $n$ is the number of the proposal bounding boxes, $x_{i}^{cls} \in \mathcal{C}$ represents the class label of the $i$-th proposal bounding box, $x_{i}^{bbox} \in \mathbb{R}^{4}$ depicts the bounding box offsets relative to the $i$-th proposal box coordinates, and $x_{i\rightarrow j} \in \mathcal{R}$ is the relationship predicate between the $i$-th and the $j$-th proposal bounding boxes. Generally, the ground-truth posterior $P(x|I,B_{I})$ is computationally intractable. Therefore, in current research, a tractable variational distribution $Q(x)$ is often used to approximate the ground-truth posterior and we need to accomplish the following MAP inference to obtain the optimal interpretation: 
\begin{equation}
x^{*}=\underset{x \in X}{argmax}P(x|I,B_{I})
\end{equation}
where $x \in X$ is a possible configuration or interpretation and $P(x|I,B_{I})$ is generally considered as the target posterior, which can also be derived as follows:
\begin{equation}
P(x|I,B_{I})=\frac{exp(-E(x,I,B_{I})}{Z(I,B_{I})}
\end{equation}
where $E(x,I,B_{I})$ is the energy function, which computes the assignment cost for the potential interpretation. $Z(I,B_{I})$ is the associated partition function. Generally, the energy function can be factorized into a summation of numerous potential terms. For instance, the following equation demonstrates one possible factorization:
\begin{equation}
E(x,I,B_{I})=\sum_{i}\psi_{u}(x_{i},I,B_{I})+\sum_{i\neq j}\psi_{b}(x_{i}, x_{j})
\end{equation}
where $\psi_{u}(x_{i},I,B_{I})$ represents the unary potential term and $\psi_{b}(x_{i}, x_{j})$ depicts the binary potential term. Essentially, the unary potential terms relate to the visual perception module, while the higher order potential terms characterize the visual context reasoning module. 

Furthermore, suppose the above energy function is parametrized as $E_{\theta}(x,I,B_{I})$, the current visual semantic information pursuit methods tend to minimize the corresponding target variational free energy $F(\theta,Q)$ by applying different deep learning-based coordinate descent strategies:
\begin{equation}
F(\theta,Q)=\sum_{x \in X}Q(x)E_{\theta}(x,I,B_{I})
\end{equation}
where the optimal $Q$ and $\theta$ can be obtained by alternating between a MAP inference step and a model selection step. In the current literature, deep learning-based message passing strategies are generally applied to implement the MAP inference step, while the model selection step is often accomplished by SGD methods. Luckily, since the message passing update rule has already implicitly accomplished the MAP inference step, it is not necessary to explicitly state the variational free energy if one choose to use message passing optimization strategy. In other words, one can choose different types of variational free energy by changing the corresponding message passing update rules.

\subsection{Training Strategy}

The existing visual semantic information pursuit methods generally follow two main training strategies: modular training and end-to-end training. Within the model selection step, the error differentials of the former one are only allowed to back-propagate within the visual context reasoning module, while the latter one can further back-propagate the error differentials to the previous visual perception module so that the whole learning system can be trained end-to-end. Essentially, within the modular training strategy, the the visual context reasoning module can be considered as a post-processing stage of the visual perception. For instance, \cite{jahangiri2017information} formulates the sequential scene parsing task as a binary tree graphical model and proposes a Bayesian framework to infer the associated target posterior. Specifically, three variants of VGG nets \cite{simonyan2014very} are applied as the visual perception module, while the proposed Bayesian framework is used as the post-processing visual context reasoning module. To accomplish the visual semantic segmentation task, \cite{chen2015semantic}, \cite{chen2018deeplab} use FCNs as the visual perception modules and apply CRF models as the post-processing visual context reasoning modules.

Instead of using modular training, the current visual semantic information pursuit methods tend to apply end-to-end training. Moreover, they tend to use deep learning based variational free energy minimization methods to model the visual context reasoning module. Such changes have several advantages: 1) since the error differentials within the visual context reasoning module can be back-propagated to the previous visual perception module, the whole system can be trained end-to-end and the final performance would be improved accordingly; 2) with the deep learning models, the classical inference operations like message passing or aggregation can be easily accomplished by a simple tensor manipulation; 3) the visual context reasoning module based on deep learning models can fully utilize the advanced parallel capability of the modern GPUs so that the inference speed can be improved. 
    
\section{major developments}

In this survey, we will limit the discussion to the key deep learning based visual context reasoning methods. Specifically, based on the manners the prior knowledge is applied, the current deep learning based visual context reasoning methods can be categorized as either bottom-up or top-down. In the following sections, we will introduce the major developments of these two directions in terms of the applied variational free energies and the corresponding optimization methods.

\subsection{Major Developments in Bottom-up Methods} 

A visual semantic information pursuit task can generally be represented in terms of associated probabilistic graphical models. For instance, MRFs or CRFs are often used to model the visual semantic segmentation tasks. Given a probabilistic graphical model, the visual context reasoning module is often formulated as a MAP estimation or variational free energy minimization. The optimization problem is generally NP-hard to resolve and we need to relax the original tight constraints and use variational-based methods to approximate the target posterior. Specifically, we firstly need to define an associated variational free energy and then try to find a corresponding optimization method to minimize it. In most cases, the applied variational free energy depends on the corresponding relaxation strategy. 

Even though numerous types of optimization methods \cite{zhang2018advances} are capable of minimizing the target variational free energy, one particular type of optimization strategy - message passing \cite{ross2011learning}, \cite{kappes2013comparative} - stands out from the competition and is widely applied in the current deep learning based visual context reasoning methods. This is because, unlike the sequential optimization methods (such as the steepest descent algorithm \cite{battiti1992first} and its many variants), the message passing strategy is capable of optimizing different decomposed sub-problems (usually from dual decomposition) in parallel. Furthermore, the message passing or aggregation operation can be easily accomplished by a simple tensor manipulation. For the bottom-up deep learning based visual context reasoning methods, only the internal prior knowledge can be used to regularize the associated variational free energies. Within the existing bottom-up methods, numerous message passing variants have been proposed in various deep learning architectures, which can be summarized as follows:

\subsubsection{Triplet-based Reasoning Models}

The visual relationship triplet $(subject,predicate,object)$ plays an important role in understanding the input image. Instead of categorizing the triplet as a whole, the current bottom-up methods tend to jointly classify each component since the computational complexity would reduce from $\mathcal{O}(N^{2}R)$ to $\mathcal{O}(N+R)$ (for $N$ objects and $R$ predicates). However, it is extremely hard to detect predicates since they often obey a long-tail distribution (the complexity become quadratic when considering all possible subject-object pairs). Given a set of object classes $\mathcal{C}$ and a set of relationship types $\mathcal{R}$, the triplet variable can be depicted as $X=\{x_{s}^{cls}, x_{s}^{bbox}, x_{o}^{cls}, x_{o}^{bbox}, x_{r}\}$, where $x_{*}^{cls} \in \mathcal{C}$ represents the class label of the associated proposal bounding box, $x_{*}^{bbox} \in \mathbb{R}^{4}$ depicts the bounding box offsets relative to the associated proposal box coordinates, and $x_{r} \in \mathcal{R}$ is the relationship predicate. The aim of the triplet-based reasoning model is to maximize the posterior $P(X|V_{S},V_{R},V_{O})$, in which $V_{S},V_{R},V_{O}$ represent the associated observed feature vectors. Through modeling the unary potential terms of the associated variational free energies, CNN-based visual perception modules are generally used to generate the above feature vectors.
\begin{figure}[!t]
\centering
\includegraphics[width=\linewidth]{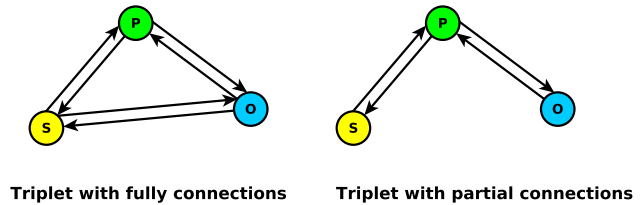}
\caption{Message passing strategies for different types of the triplet-based reasoning models, in which the messages are passing among the corresponding triplet components.}
\end{figure} 

In the current literature, the triplet-based reasoning models generally use CNN architectures to model higher order potential terms. Moreover, to minimize the variational free energy, the associated message passing strategies are often proposed based on the connection configuration of the triplet graphical model, as demonstrated in Fig.4. One typical triplet-based reasoning model is the relationship proposal network \cite{zhang2017relationship}, which formulates the triplet structure as a fully connected clique and only considers a third-order potential term within the associated variational free energy. Two CNN-based compatibility modules are proposed to model the third-order potential terms so that a consistent unary prediction combination is rendered more likely. Inspired by the faster RCNN model, the authors in \cite{li2017vip} propose a CNN-based phrase-guided message passing structure (PMPS) to infer input triplet proposals, in which the subjects and the objects are only connected through the predicates. They place the predicate at the dominant position and specifically design a gather-and-broadcast message passing strategy, which is applied in both convolutional and fully connected layers. Unlike the above methods, \cite{zhang2017visual} proposes to model the predicate as a vector translation between the subject and the object, in which both subject and object are mapped into a low-dimensional relation space with less variance. Instead of using the conventional ROI pooling, the authors use the bilinear feature interpolation to transfer knowledge between object and predicate.

\subsubsection{MRF-based or CRF-based Reasoning Models}

MRFs and CRFs are commonly used undirected probabilistic graphical models in computer vision community. They are capable of capturing rich contextual information exhibited in natural images or videos. MRFs are generative models while CRFs are discriminative models. Most of the visual semantic information pursuit applications, in particular the visual semantic segmentation tasks, are often formulated as MRFs or CRFs. Given an input stimuli $I$ and the variable $X$ representing the semantic information of interest, the aim of the MRF-based or CRF-based reasoning model is to maximize the posterior $P(X|I)$ or minimize the variational free energy $F(Q)$. The exact inference methods only exist for special MRF or CRF structures such as conjunction trees or local cliques. For instance, the authors in \cite{dai2017detecting} propose a deep relational network to detect the potential visual relationship triplets. They apply CRFs to model the associated fully connected triplet cliques and use the sequential CNN computing layers to exactly infer the corresponding marginals factorized from the joint posterior.
\begin{figure}[!t]
\centering
\includegraphics[width=1.6 in]{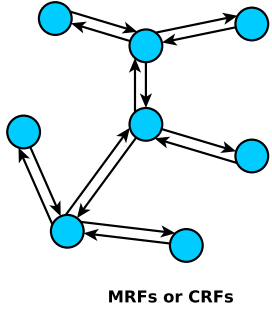}
\caption{Message passing strategies for MRF-based or CRF-based reasoning models, in which the messages are generally passing within the same semantic level.}
\end{figure} 
 
In the current literature relating to the bottom-up approach, the general MRF or CRF structures often need variational inference methods such as mean field (MF) approximation \cite{georges1996dynamical}, \cite{barabasi1999mean} or loopy belief propagation (BP) \cite{murphy1999loopy} to infer the target posterior. Specifically, the applied variational free energy is often devised depending on the relaxation strategy of the corresponding constraint optimization problem, and the message passing optimization methodology is generally applied to minimize the above variational free energy. Fig.5 shows the general message passing strategies of the MRF-based or CRF-based reasoning models. A well-known CRF-based reasoning model is the CRF-as-RNN \cite{zheng2015conditional}, \cite{arnab2018conditional}, which incorporates the RNN-based visual context reasoning module in the FCN visual perception module so that the proposed visual semantic segmentation system can be trained end-to-end. Specifically, FCN layers are used to formulate the unary potentials of the DenseCRF model \cite{krahenbuhl2011efficient} while the binary potentials are formed by a sequence of CNN layers. As a result, the associated mean field inference method can be accomplished by the corresponding RNN. Essentially, two relaxation measures are applied within the mean field approximation: 1) the tractable variational distribution $Q(X)$ is used to approximate the underling posterior $P(X|I)$; 2) the joint variational distribution can be fully factorized into a combination of independent nodes, which implies maximizing the marginal of each independent node is guaranteed to accomplish the original MAP estimation. Inspired by the above methodology, the authors in \cite{xu2017scene} use a more general RNN architecture - gated recurrent units (GRUs) - to generate the scene graphs from the input images using the mean filed inference method. Specifically, they use the internal memory cells in GRUs to store the generated contextual information and apply a primal-dual update rule to speed up the inference procedure. Unlike the above methods that optimize the CRFs using iterative strategy, the deep parsing network (DPN) proposed in \cite{liu2017deep} is able to achieve high visual semantic segmentation performance by only applying one iteration of MF, which also can be considered as a generalized case of the existing models since it can represent various types of binary potential terms. 
\begin{figure}[!t]
\centering
\includegraphics[width=3.0 in]{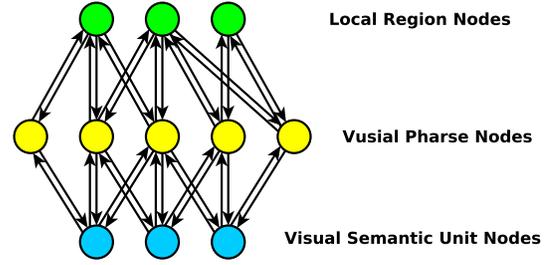}
\caption{Message passing strategies for the visual semantic hierarchy reasoning models, in which the messages are passing though different semantic levels.}
\end{figure} 

\subsubsection{Visual Semantic Hierarchy Reasoning Models }

In visual semantic information pursuit tasks, visual semantic hierarchies are ubiquitous. They often consist of visual semantic units, visual semantic phrases, local visual semantic regions and the scene graph. Given a visual semantic hierarchy, the contextual information within other semantic layers are often used to maximize the posterior of the current semantic layer. Essentially, such inference procedure tend to have a tighter upper bound for the target variational free energy and thus often obtains a better MAP inference performance. Specifically, to dynamically build visual semantic hierarchies, the visual semantic hierarchy reasoning models are often required to align the contextual information of different visual semantic levels. As shown in Fig.6, through passing the contextual information among different visual semantic levels within the generated visual semantic hierarchy, each visual semantic level can obtain much more consistent posterior.

Unlike the previous methods that only model pairwise potential terms within the same visual semantic level, the structure inference network (SIN) \cite{liu2018structure} propagates contextual information from the holistic scene and the adjacent connected nodes to the target node. Within this object detection framework, GRUs are used to store the contextual information. Motivated by the computational consideration, the maximum pooling layer is applied to aggregate messages from the adjacent connected nodes into an integrated message. To leverage the contextual information across different semantic levels, multi-level scene description network (MSDN) \cite{li2017scene} establishes a dynamic graph consisting of object nodes, phrase nodes and region nodes. For each semantic level, a CNN-based merge-and-refine strategy is proposed to pass the contextual information along the graph structure.

\subsubsection{DAG-based Reasoning Models}

Generally, the variational free energy employed in CRFs usually fails to enforce higher-order contextual consistency due to the computational considerations. Furthermore, small-sized objects are often smoothed out by the CRFs, which degrades the semantic segmentation performance. Instead of applying conventional CRFs, some bottom-up methods tend to use undirected cyclic graphical (UCG) models to formulate the visual semantic segmentation tasks. However, due to its loopy property, UCG models generally cannot be formulated as RNNs. To resolve this issue, as shown in Fig.7, UCGs are often decomposed into a sequence of directed acyclic graphs (DAGs), in which each DAG can be modelled by a corresponding RNN. Such a RNN architecture is also known as DAG-RNN, which explicitly propagate local contextual information based on the directed graphical structure. Essentially, a DAG-based reasoning model like the DAG-RNN has two main advantages: 1) compared with conventional CNN models (such as FCNs), it is empirically found to be significantly more effective at aggregating context; 2) it requires substantially less parameters as well as demanding fewer computation operations, which makes it more favourable for applications on resource-limited embedded platforms.
\begin{figure}[!t]
\centering
\includegraphics[width=\linewidth]{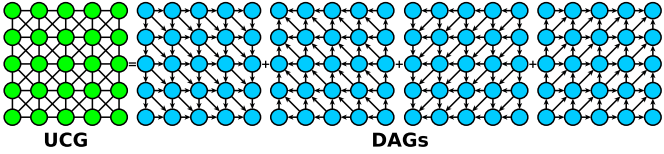}
\caption{The applied UCG is decomposed as a sequence of DAGs (one possible decomposition), in which the messages in each DAG are passing along its specific structure.}
\end{figure} 

In current DAG-based reasoning models \cite{shuai2018scene}, \cite{shuai2016dag}, the DAG-RNNs apply 8-neighborhood UCG graphs to effectively encode the long-range contextual information so that the discriminative capabilities of the local representations are greatly improved. Inspired by the conventional tree-reweighed max-product algorithm (TRW) \cite{wainwright2005new}, the applied UCGs are decomposed into a sequence of DAGs, in which any vertex pair can be mutually reachable in the resulting set of DAGs. Furthermore, the DAG-RNNs are often integrated with convolution and deconvolution layers, and a novel class-weighted loss is applied since the class occurrence frequencies are generally imbalanced in most visual semantic information pursuit tasks, especially the visual semantic segmentation applications.

\subsubsection{External Memory Reasoning Models}

One of difficult issues for the visual semantic information pursuit is to resolve the dataset imbalance problem. To address this issue, one generally needs to accomplish the so-called few-shot learning tasks \cite{socher2013zero}, \cite{romera2015embarrassingly}, \cite{vinyals2016matching} since most categories in the datasets have only few training samples. Unlike the above models, instead of using the internal memory cells like long-short term memory (LSTM) or gated recurrent unit (GRU), the external memory reasoning models apply the external memory cells, such as neural turing machine (NTM) \cite{graves2014neural} or memory-augmented neural network (MANN) \cite{santoro2016meta}, to store the generated contextual information. More importantly, they tend to use meta-learning strategy \cite{santoro2016meta} within the inference procedure. Such a meta-learning strategy can be summarized as "learning to learn", which selects parameters $\theta$ to reduce the expected learning cost $\mathcal{L}$ across a distribution of datasets $p(D)$: $\theta^{*}=argmin_{\theta}E_{D\sim p(D)}[\mathcal{L}(D;\theta)]$. To prevent the network from slowly learning specific sample-class bindings, it directly stores the new input stimuli at the corresponding external memory cells instead of relearning them. Through such meta-learning, the convergence speed of the visual semantic information pursuit task is greatly improved so that only few training samples are enough to converge at a stable status.
\begin{figure}[!t]
\centering
\includegraphics[width=3.2 in]{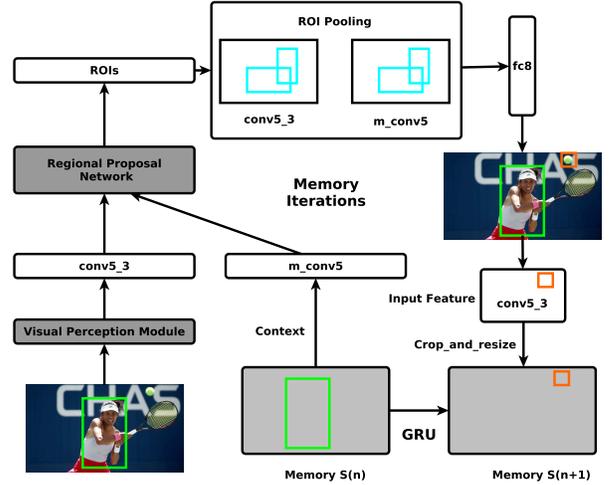}
\caption{Overview of 2-D spatial external memory iterations for object detection. The old detection is marked with a green box, and the new detection is marked with orange. Here, the spatial memory network is only unrolled one iteration.}
\end{figure}

Instead of detecting objects in parallel like the conventional object detection methods, the authors in \cite{chen2017spatial} propose a novel instance-level spatial reasoning strategy, which tries to recognize objects conditioned on the previous detections. To this end, spatial memory network (SMN) \cite{chen2017spatial}, a 2-D spatial external memory, is devised to store the generated contextual information and extract spatial patterns by using a effective reasoning module. Essentially, this leads to a new sequential reasoning module where image and memory are processed in parallel to obtain detections which update the memory again, as shown in Fig.8. Unlike the above method that made sequential updates to memory, the authors in \cite{chen2018iterative} propose to update the regions in parallel as an approximation, in which a cell can be covered multiple times from different regions in overlapping cases. Specifically, a weight matrix is devised to keep track of how much a region has contributed to a memory cell. The final value of each updated cell is the weighted average of all regions.

\subsection{Major Developments in Top-down Methods}
 
For visual semantic information pursuit applications, the associated visual semantic relations generally reside in a huge semantic space. Unfortunately, only limited training samples are available, which implies it is impossible to fully train every possible visual relation. To maximize the target posterior from this long-tail distribution, the existing top-down methods generally transform the MAP inference tasks into linear programming problems. More importantly, they often distill the external linguistic prior knowledge into the associated learning systems so that the objective functions of the target constraint optimization problems can be further regularized accordingly. Therefore, compared with the bottom-up methods, the top-down methods generally converge relatively easily. In this section, based on their distillation strategies, we divide the existing top-down methods into the following categories:
\begin{figure}[!t]
\centering
\includegraphics[width=\linewidth]{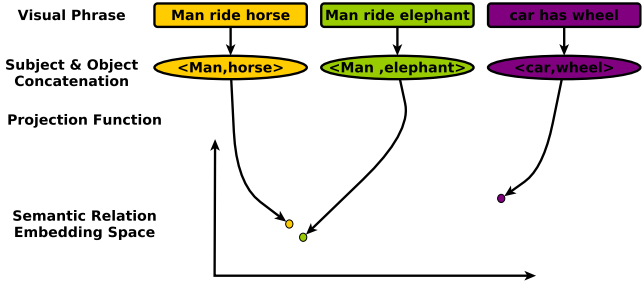}
\caption{The diagram of semantic affinity distillation models.}
\end{figure} 

\subsubsection{Semantic Affinity Distillation Models}

Even though the visual semantic relations obey a long-tail distribution, they are often semantically related to each other, which means it is possible to infer an infrequent relation from similar relations. To this end, the semantic affinity distillation models project the corresponding feature vectors (which are often generated from the union of bounding boxes of the associated objects) into a low-dimensional semantic relation embedding space and use their semantic affinities as the external prior knowledge to regularize the target optimization problem. In general, the projection function would be trained by enforcing similar visual semantic relations to be close together in the semantic relation embedding space. For instance, the visual semantic relation $(man-ride-horse)$ should be close to $(man-ride-elephant)$ and far away from $(car-has-wheel)$ in the associated semantic relation embedding space, as illustrated in Fig.9. The semantic affinity distillation models are capable of resolving zero-shot learning tasks since the visual semantic relations without any training samples can still be recognized by the external linguistic knowledge, which is clearly impossible for the bottom-up methods that only use internal visual prior knowledge.

One of the pioneering works is the visual relationship detection with language prior method \cite{lu2016visual}, which trains the visual models for objects and predicates individually, and later combines them together by applying the external semantic affinity-based linguistic knowledge to predict consistent visual semantic relations. Unlike the above algorithm, the context-aware visual relationship detection method \cite{zhuang2017towards} tries to recognize the predicate by incorporating the subject-object pair semantic contextual information. Specifically, the context is encoded via word2vec into a semantic embedding space and is applied to generate a classification result for the predicate. To summarize, the external semantic affinity linguistic knowledge can not only improve the inference speed but also leads to zero-shot generalizations.
\begin{figure}[!t]
\centering
\includegraphics[width=\linewidth]{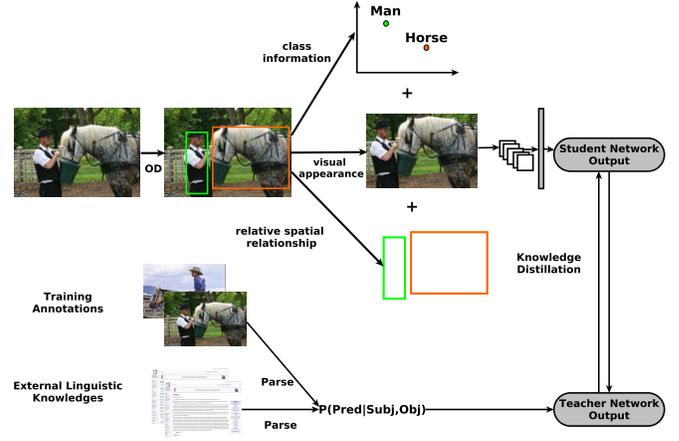}
\caption{The diagram of teacher-student distillation models.}
\end{figure}

\subsubsection{Teacher-student Distillation Models}

To resolve the long-tail distribution issue, instead of relying on semantic affinities, the teacher-student distillation models tend to use external linguistic knowledge (the conditional distribution of a visual semantic relation given specific visual semantic units) generated from the public knowledge databases to constrain the target optimization problem. Given the input stimuli $X$ and the predictions $Y$, the optimal teacher network $t$ is selected from an associated candidate set $T$ by minimizing $KL(t(Y)||s_{\phi}(Y|X))-C\mathbb{E}_{t}[L(X,Y)]$, where $t(Y)$ and $s_{\phi}(Y|X)$ represent the prediction results of the teacher and student networks, respectively; $\phi$ is the parameter set of the student network and $C$ is a balancing term; $L(X,Y)$ depicts the constraint function, in which the predictions that satisfy the constraints are rewarded and the remaining are penalized; $KL$ measures the $KL$ divergence of the teacher's and the student's prediction distributions. Essentially, through solving the above optimization, the teacher's output can be viewed as a projection of the student's output in the feasible polytopes constrained by the external linguistic prior knowledge. 

However, the teacher network itself, in most cases, is not enough to provide accurate predictions since the external linguistic prior knowledge is often noisy. In general, the student network represents the architecture without any external linguistic knowledge, while the framework incorporating both internal visual and external linguistic knowledge is formulated as the teacher network. They each have their own advantages: the teacher outperforms in cases with sufficient training samples, while the student achieves superior performance in few-shot or zero-shot learning scenarios. Therefore, unlike the previous distillation methods \cite{hinton2015distilling}, \cite{hu2016harnessing}, \cite{hu2016deep} that only use either the teacher or the student as the output, the current teacher-student distillation models \cite{yu2017visual}, \cite{plesse2018visual} tend to incorporate the prediction results from both student and teacher networks, as shown in Fig.10.

\section{future research directions}

Even though the current visual semantic information pursuit methods have achieved satisfying performance not seen before, there are still numerous challenging yet exciting research directions to investigate in the future. 

\subsection{Weakly-supervised Pursuit Methods}

Annotations for the visual semantic information pursuit applications, especially the visual semantic segmentation tasks, are generally hard to obtain since we need to invest tremendous times and efforts into the labelling procedure. However, most current visual semantic information pursuit methods are essentially fully-supervised algorithms, which implies rich annotations are required if we want to train these methods. For instance, to locate the objects, the fully-supervised methods require ground-truth locations of the associated bounding boxes. To alleviate the annotation burden, the weakly-supervised pursuit methods \cite{bilen2015weakly}, \cite{bilen2016weakly}, \cite{wei2017stc}, \cite{cinbis2017weakly}, \cite{zhang2018leveraging} have been proposed in recent years, which only require the vocabulary information to extract the visual semantic information. Unfortunately, even though the current weakly-supervised pursuit methods generally require more computation, they still do not achieve comparable performance to fully-supervised pursuit methods. This is the reason why the fully-supervised methods are prevalent in the visual semantic information pursuit literature.

\subsection{Pursuit Methods using Region-based Decomposition}

For most visual semantic information graphical models, it is generally computationally intractable to infer the target posteriors. To resolve this NP-hard problem, mean field approximation is often applied in the current visual semantic information pursuit methods, in which the associated graphical model is fully decomposed into independent nodes. Unfortunately, such simple decomposition strategy only incorporates unary pseudo-marginals into the associated variational free energy, which is clearly not enough for the complicated visual semantic information pursuit applications. Moreover, for the generated dense (fully connected) inference model, it is often slow to converge. Inspired by the generalized belief propagation algorithm \cite{yedidia2001generalized}, there are pursuit methods \cite{li2018factorizable}, \cite{herzig2018mapping} starting to apply region-based decomposition strategy, in which the associated graphical model is factorized into various regions and the nodes within each region are not independent. The applied region-based decomposition strategy can not only improve the inference speed in some cases, but also incorporate higher-order pseudo-marginals into the associated variational free energy. However, there are still several open questions: 1) How many regions are enough for most visual semantic information pursuit applications? 2) How to efficiently compute the higher-order pseudo-marginals given the decomposed regions? 3) How to properly propagate contextual information between different regions?

\subsection{Pursuit Methods with Higher-order Potential Terms}

To extract the visual semantic information, the scene modelling methods generally need to factorize the associated energy functions into various potential terms, in which the unary terms produce the predictions while the higher-order potential terms constrain the generated predictions to be consistent. However, the current visual semantic information pursuit methods typically incorporate only pair-wise potential terms into the associated variation free energy, which is clearly not enough. To resolve this issue, the current visual semantic information pursuit methods start to incorporate higher-order potential terms into the associated variational free energies. For instance, for the visual semantic segmentation tasks, the recently proposed UCG-based pursuit methods \cite{shuai2018scene}, \cite{shuai2016dag} replace the pair-wise potential terms (applied in most CRF-based models) with higher potential terms, and thus achieve the state-of-the-art segmentation performance. However, through incorporating the higher-order potential terms, the target constraint optimization problems become much harder to resolve since the polynomial higher-order potential terms would inject more non-convexities into the objective function \cite{fazelnia2018crvi}. Therefore, further efforts are needed to address this non-convexity issue.

\subsection{Pursuit Methods with Advanced Domain Adaptation}

One of the most difficult issues for the visual semantic information pursuit applications is the dataset imbalance problem. In most cases, only few categories have enough training samples, while the remaining either have few or even zero training samples. Due to the long-tail distribution, such situation would become even worse when we try to pursue the visual semantic relation information. To address this few-shot or zero-shot learning issue, domain adaptation \cite{glorot2011domain}, \cite{pan2011domain}, \cite{ganin2015unsupervised}, \cite{patel2015visual} becomes a natural choice since it can transfer the missing knowledge from the related domains into the target domain. For instance, to resolve the corresponding few-shot learning problems, the top-down pursuit methods use the distilled external linguistic knowledge to regularize the variational free energy, while the bottom-up pursuit methods achieve meta-learning through using the external memory cells. Essentially, the current domain adaptation strategies used in existing visual semantic information pursuit methods mainly focus on learning generic feature vectors from one domain that are transferable to other domains. Unfortunately, they generally transfer unary features and largely ignore more structured graphical representations \cite{yang2018glomo}. To transfer the structured graphs to the corresponding domains, more advanced domain adaptation methodologies are much needed in the future.

\subsection{Pursuit Methods without Message Passing}

To resolve the target NP-hard constrained optimization problems, even though numerous constraint optimization strategies are available \cite{kappes2013comparative}, the current deep learning based visual semantic information pursuit methods totally depend on one specific optimization methodology - message passing. Essentially, message passing is generally motivated by linear programming and variational optimization. It is proven that, for modern deep learning architectures, such parallel optimization methodology is more effective than other optimization strategies. Besides being widely applied in different visual semantic information pursuit tasks, it also succeeds in the quantum chemistry area \cite{li2016gated}, \cite{battaglia2016interaction}, \cite{schutt2017quantum}, \cite{gilmer2017neural}. However, the parallel message passing strategies empirically underperform compared to the sequential optimization methodologies and typically do not provide feasible integer solutions \cite{kappes2013comparative}. To address these issues, the visual semantic information pursuit methods without using message passing optimization strategy are certainly needed in the future.

\section{benchmarks and evaluation metrics}

In this section, we will introduce the main benchmarks and evaluation metrics for the four research applications investigated in this survey, which include object detection, visual semantic segmentation, visual relationship detection and scene graph generation.

\subsection{Object Detection}
\subsubsection{Benchmarks}

Two benchmarks are commonly used in object detection applications: the test set of PASCAL VOC 2007 \cite{everingham2007pascal} and the validation set of MS COCO \cite{lin2014microsoft}. More specifically, the test set of PASCAL VOC 2007 contains 4,952 images and 14,976 object instances from 20 categories. To evaluate the performances of the object detection methods in different image scenes, it includes large number of objects with abundant variations within each category, viewpoint, scale, position, occlusion and illumination. Compared with PASCAL VOC 2007 test set, the MS COCO benchmark is more challenging since the images in this dataset are gathered from complicated day-to-day scenes that contain common objects in their natural contexts. Specifically, MS COCO benchmark contains 80,000 training images and 500,000 instance annotations. To evaluate the detection performance, most object detection methods often use the first 5,000 MS COCO validation images. Additional non-overlapping 5,000 images, in some cases, have also been used as the validation dataset.

\subsubsection{Evaluation Metrics}

To evaluate the object detection methods, one needs to consider the following two performance measures: the object proposals generated by object detection methods and the corresponding objectness detection. In the existing literatures, the metrics for evaluating object proposals are often functions of intersection over union (IOU) between the proposal locations and the associated ground-truth annotations. Given the IOU, recall can be obtained as the fraction of ground-truth bounding boxes covered by proposal locations above a certain IOU overlap threshold. To evaluate the performance of the objectness detection, mean average precision (mAP) metric is often used for VOC 2007 test benchmark (the IOU threshold is normally set to $0.5$), while MS COCO 2015 test-dev benchmark generally apply two types of metrics: average precision (AP) over all categories and different IOU thresholds, and average recall (AR) over all categories and IoUs (which is basically computed on a per-category basis, i.e. the maximum recall given a fixed number of detections per image). Specifically, $AP$, $AP^{50}$, $AP^{70}$ represent the average precision over different IOU thresholds from $0.5$ to $0.95$ with a step of $0.05$ (written as 0.5:0.95), the average precision with IOU threshold of $0.5$ and the average precision with IOU threshold of $0.7$, respectively. $AR^{1}$, $AR^{10}$, $AR^{100}$ depict the average recall given $1$, $10$, $100$ detections per image, respectively. We recommend interested readers refer to the relevant papers \cite{ren2015faster}, \cite{redmon2017yolo9000} for the details and the mathematical formulations of the above metrics.

\subsection{Visual Semantic Segmentation}
\subsubsection{Benchmarks}

In this survey, three main benchmarks are chosen out from the abundant datasets for visual semantic segmentation methods: \textit{Pascal Context} \cite{mottaghi2014role}, \textit{Sift Flow} \cite{liu2009nonparametric} and \textit{COCO Stuff} \cite{caesar2016coco}. Specifically, \textit{Pascal Context} benchmark contains 10,103 images extracted from the Pascal VOC 2010 dataset, in which 4,998 images are used for training. The images are relabelled as pixel-wise segmentation maps which include 540 semantic categories (including the original 20 categories) and each image has approximately the size of $375 \times 500$. \textit{Sift Flow} dataset contains 2,688 images obtained from 8 specific kinds of outdoor scenes. Each image has the size of $256 \times 256$, which belongs to one of the 33 semantic classes. \textit{COCO Stuff} is a recently released scene segmentation dataset. It includes 10,000 images extracted from the Microsoft COCO dataset, in which 9,000 images are used for training and the previous unlabelled stuff pixels are further densely annotated with extra 91 classes. 

\subsubsection{Evaluation Metrics}

To evaluate the performances of the visual semantic segmentation methods, three main metrics are generally applied in the existing literatures: \textit{Global Pixel Accuracy} (GPA), \textit{Average per-Class Accuracy} (ACA) and \textit{mean Intersection of Union} (mIOU). Specifically, GPA represents the percentage of all correctly classified pixels, ACA depicts the mean of class-wise pixel accuracy and mIOU is the mean of the accuracy metric IOU. The details and the corresponding mathematical formulations of  the above metrics can be found in \cite{long2015fully}. 

\subsection{Visual Relationship Detection}
\subsubsection{Benchmarks}

The current visual relationship detection methods often use two benchmarks: \textit{visual relationship dataset} \cite{lu2016visual} and \textit{visual genome} \cite{krishna2017visual}. Unlike the datasets for object detection, the visual relationship datasets should contain more than just objects localized in the image. Instead, they should capture the rich variety of interactions between the subject and the object pairs. Various types of interactions are considered in the above visual relationship benchmark datasets, i.e. verbs (e.g. wear), spatial (e.g. in front of), prepositions (e.g. with) or comparative (e.g. higher than). Moreover, the types of predicates per category should be large enough. For instance, a $man$ can be associated with the predicates such as $wear$, $with$, $play$, etc. Specifically, visual relationship dataset contains 5000 images with 100 object categories and 70 predicates. In total, the dataset contains 37,993 relationships with 6,672 relationship types and 24.25 predicates per object category. Unlike the visual relationship dataset, the recently proposed visual genome dataset incorporates numerous kinds of annotations, one of which is visual relationships. The visual genome relationship dataset contains 108,077 images and 1,531,448 relationships. However, it generally needs to be cleansed since the corresponding annotations often contain some misspellings and noisy characters, and the verbs and the nouns are also in different forms.

\subsubsection{Evaluation Metrics}

The current visual relationship detection methods often use two evaluation metrics: $recall @ 50$ and $recall @ 100$. Here, $recall @ x$ \cite{alexe2012measuring} represents the fraction of times the correct relationship is predicted in the top $x$ confident relationship predictions. The reason why we use $recall @ x$ instead of widely applied mean average precision (mAP) metric is because mAP is a pessimistic evaluation metric, meaning we can not exhaustively annotate all possible relationships in an image. Even if the prediction is correct, mAP still would penalize the prediction if we do not have that particular ground truth annotation.

\subsection{Scene Graph Generation}
\subsubsection{Benchmarks}

The \textit{visual genome} \cite{krishna2017visual} is often used as the benchmark for the scene graph generation applications. Unlike the previous visual relationship datasets, the visual relationships within the visual genome scene graph dataset locate in the associated scene graphs and are generally not independent of each other. Specifically, the visual genome scene graph dataset contains 108,077 images with an average of 38 objects and 22 relationships per image. However, a substantial fraction of the object annotations have poor quality and overlapping bounding boxes and/or ambiguous object names. Therefore, a clean visual genome scene graph generation dataset is often needed in the real evaluation procedure. In the current scene graph generation methods, instead of training on all possible categories and predicates, the most frequent categories and predicates are often chosen for evaluation.

\subsubsection{Evaluation Metrics}

Similar to the previous visual relationship detection methods, the scene graph generation methods generally apply $recall @ x$ \cite{alexe2012measuring} metric instead of the mAP metric. Specifically, $recall @ 50$ and $recall @ 100$ are often used to evaluate the corresponding scene graph generation methods.

\section{experimental comparison}

In this section, we compare the performance of different visual semantic information pursuit methods for each potential application mentioned in this survey. Specifically, for each of the following subsection, we will choose the most representative methods to compare the pursuit performance. Moreover, the benchmarks and the evaluation metrics mentioned in the above section will be used to accomplish the performance comparisons.

\subsection{Object Detection}

In this section, two benchmarks - VOC 2007 test \cite{everingham2007pascal} and MS COCO 2015 test-dev \cite{lin2014microsoft} - are applied to compare different cutting-edge object detection methods. For VOC 2007 test dataset, we select 5 current object detection methods including Fast R-CNN \cite{girshick2015fast}, Faster R-CNN \cite{ren2015faster}, SSD500 \cite{liu2016ssd}, ION \cite{bell2016inside} and SIN \cite{liu2018structure}, as shown in Table \ref{od_1}; For MS COCO 2015 test-dev benchmark, Fast R-CNN \cite{girshick2015fast}, Faster R-CNN \cite{ren2015faster}, YOLOv2 \cite{redmon2017yolo9000}, ION \cite{bell2016inside} and SIN \cite{liu2018structure} are included in the performance comparison task shown in Table \ref{od_2}. Among the above methods, only ION \cite{bell2016inside} and SIN \cite{liu2018structure} incorporate the visual context reasoning modules within the learning procedure while others merely apply visual perception modules. The reason for incorporating various state-of-the-art visual perception models in the comparison is to provide a complete comparison and to gain further understanding of the impact of the visual context reasoning modules.
\begin{table}[!t]
   \centering
   \begin{threeparttable}
	\renewcommand{\arraystretch}{1.5}
	\caption{Performance comparison on VOC 2007 test.}
	\label{od_1}
	\centering
    \begin{tabularx}{\columnwidth}{*3Y}
	\toprule
	{Method} & {Train} & {$mAP$}\\
	\midrule
	Fast R-CNN \cite{girshick2015fast} & $07+12$ & $70.0$\\
	Faster R-CNN \cite{ren2015faster} & $07+12$ & $73.2$\\
	SSD500 \cite{liu2016ssd} & $07+12$ & $75.1$\\
	ION \cite{bell2016inside} & $07+12$ & $75.6$\\
	SIN \cite{liu2018structure} & $07+12$ & $\mathbf{76.0}$\\
	\bottomrule
    \end{tabularx}
      \begin{tablenotes}
	\item [\textbullet] Note: $07+12$ represents 07 trainval + 12 trainval.
      \end{tablenotes}
    \end{threeparttable}
\end{table} 

\begin{table}[!t]
   \resizebox{\columnwidth}{!}{
   \centering
   \begin{threeparttable}
	\renewcommand{\arraystretch}{1.5}
	\caption{Performance comparison on COCO 2015 test-dev.}
	\label{od_2}
	\centering
        \begin{tabular}{*8c}
	\toprule
	{Method} & {Train} & {$AP$} & {$AP^{50}$} & {$AP^{70}$} & {$AR^{1}$} & {$AR^{10}$} & {$AR^{100}$}\\
	\midrule
	Fast R-CNN \cite{girshick2015fast} & $train$ & $20.5$ & $39.9$ & $19.4$ & $21.3$ & $29.5$ & $30.1$\\
	Faster R-CNN \cite{ren2015faster} & $train$ & $21.1$ & $40.9$ & $19.9$ & $21.5$ & $30.4$ & $30.8$\\
	YOLOv2 \cite{redmon2017yolo9000} & $trainval35k$ & $21.6$ & $44.0$ & $19.2$ & $20.7$ & $31.6$ & $\mathbf{33.3}$\\
	ION \cite{bell2016inside} & $train$ & $23.0$ & $42.0$ & $\mathbf{23.0}$ & $\mathbf{23.0}$ & $\mathbf{32.4}$ & $33.0$\\
	SIN \cite{liu2018structure} & $train$ & $\mathbf{23.2}$ & $\mathbf{44.5}$ & $22.0$ & $22.6$ & $31.6$ & $32.0$\\
	\bottomrule
      \end{tabular}
      \begin{tablenotes}
	\item [\textbullet] Note: $trainval35k$ represents COCO train + 35k val \cite{bell2016inside}.
      \end{tablenotes}
    \end{threeparttable}
    }
\end{table}

In Table \ref{od_1} and \ref{od_2}, we can observe that the object detection methods with the visual context reasoning modules (such as ION and SIN) can generally achieve better performance than the current visual perception object detection algorithms (such as Fast R-CNN, Faster R-CNN, SSD500 and YOLOv2). This is because they consider the object detection task as a combination of perception and reasoning instead of only concentrating on the perception. Specifically, they consider the previous detected objects or the holistic scene as contextual information and try to improve the detection performance by inferencing over the above contextual information. In some cases, such contextual information can be quite important for detecting the target objects. For instance, when the target object is partly occluded by other objects or the target object only occupies extremely small region within the image. For such scenarios, it is almost impossible for the current visual perception object detection methods to detect the target objects. However, given the contextual information around the target objects, it is still possible to infer the target objects even in such harsh scenarios.

\subsection{Visual Semantic Segmentation} 

In this section, we compare several state-of-the-art visual semantic segmentation methods on three benchmarks: \textit{Pascal Context} \cite{mottaghi2014role}, \textit{Sift Flow} \cite{liu2009nonparametric} and \textit{COCO Stuff} \cite{caesar2016coco}. For \textit{Pascal Context}, only the most frequent 59 classes are selected for evaluation. The classes whose frequencies are lower than $0.01$ are considered as rare classes according to the 85-15 percent rule; For \textit{Sift Flow}, similar to paper \cite{liu2009nonparametric}, we split the whole dataset into training and testing sets with 2,488 and 200 images, respectively. Each pixel within the above images can be classified as one of the most frequent 33 semantic categories. Based on the 85-15 percent rule, the classes whose frequencies are lower than $0.05$ are considered as rare classes; For \textit{COCO Stuff}, each pixel can be categorized as one of 171 semantic classes in total and the frequency threshold $0.4$ is used to determine the rare classes.
\begin{table}[!t]
   \centering
   \begin{threeparttable}
	\renewcommand{\arraystretch}{1.5}
	\caption{Performance comparison (\%) on Pascal Context dataset (59 classes).}
	\label{vss_1}
	\centering
    \begin{tabularx}{\columnwidth}{bsss}
	\toprule
	{Method} & {GPA} & {ACA} & {mIOU}\\
	\midrule
	CFM \cite{dai2015convolutional} & $-$ & $-$ & $31.5$\\
	DeepLab \cite{chen2015semantic} & $-$ & $-$ & $37.6$\\
	FCN-8s \cite{shelhamer2017fully} & $67.5$ & $52.3$ & $39.1$\\
	CRF-RNN \cite{zheng2015conditional} & $-$ & $-$ & $39.3$\\
	DeepLab + CRF \cite{chen2015semantic} & $-$ & $-$ & $39.6$\\
	ParseNet \cite{liu2017deep} & $-$ & $-$ & $40.4$\\
	ConvPP-8s \cite{xie2016top} & $-$ & $-$ & $41.0$\\
	UoA-Context + CRF \cite{lin2016efficient} & $71.5$ & $53.9$ & $43.3$\\
	DAG-RNN \cite{shuai2018scene} & $72.7$ & $55.3$ & $42.6$\\
	DAG-RNN + CRF \cite{shuai2018scene} & $\mathbf{73.6}$ & $\mathbf{55.8}$ & $\mathbf{43.7}$\\
	\bottomrule
      \end{tabularx}
      \begin{tablenotes}
	\item [\textbullet] Note: For fair comparison, all the above methods apply VGG-16 \cite{simonyan2014very} as visual perception module.
      \end{tablenotes}
    \end{threeparttable}
\end{table} 

Specifically, 10 current visual semantic segmentation methods including CFM \cite{dai2015convolutional}, DeepLab \cite{chen2015semantic}, DeepLab + CRF \cite{chen2015semantic}, FCN-8s \cite{shelhamer2017fully}, CRF-RNN \cite{zheng2015conditional}, ParseNet \cite{liu2017deep}, ConvPP-8s \cite{xie2016top}, UoA-Context + CRF \cite{lin2016efficient}, DAG-RNN \cite{shuai2018scene} and DAG-RNN +CRF \cite{shuai2018scene} are used to compare on \textit{Pascal Context} benchmark, as shown in Table \ref{vss_1}. For the \textit{Sift Flow} dataset, besides the state-of-the-art semantic segmentation methods such as ParseNet \cite{liu2017deep}, ConvPP-8s \cite{xie2016top}, FCN-8s \cite{shelhamer2017fully}, UoA-Context + CRF \cite{lin2016efficient}, DAG-RNN \cite{shuai2018scene} and DAG-RNN +CRF \cite{shuai2018scene}, we also compare various previous methods like Byeon et al. \cite{byeon2015scene}, Liu et al. \cite{liu2009nonparametric}, Pinheiro et al. \cite{pinheiro2014recurrent}, Farabet et al. \cite{farabet2013learning}, Tighe et al. \cite{tighe2013finding}, Sharma et al. \cite{sharma2014recursive}, Yang et al. \cite{yang2014context} and Shuai et al. \cite{shuai2016scene}, as shown in Table \ref{vss_2}. For the recently released \textit{COCO Stuff} benchmark, we compare 5 different visual semantic segmentation methods, which include FCN \cite{caesar2016coco}, DeepLab \cite{chen2015semantic}, FCN-8s \cite{shelhamer2017fully}, DAG-RNN \cite{shuai2018scene} and DAG-RNN +CRF \cite{shuai2018scene}, as depicted in Table \ref{vss_3}.
\begin{table}[!t]
   \centering
   \begin{threeparttable}
	\renewcommand{\arraystretch}{1.5}
	\caption{Performance comparison (\%) on Sift Flow dataset (33 classes).}
	\label{vss_2}
	\centering
    \begin{tabularx}{\columnwidth}{bsss}
	\toprule
	{Method} & {GPA} & {ACA} & {mIOU}\\
	\midrule
	Byeon et al. \cite{byeon2015scene} & $70.1$ & $22.6$ & $-$\\
	Liu et al. \cite{liu2009nonparametric} & $74.8$ & $-$ & $-$\\
	Pinheiro et al. \cite{pinheiro2014recurrent} & $77.7$ & $29.8$ & $-$\\
	Farabet et al. \cite{farabet2013learning} & $78.5$ & $29.4$ & $-$\\
	Tighe et al. \cite{tighe2013finding} & $79.2$ & $39.2$ & $-$\\
	Sharma et al. \cite{sharma2014recursive} & $79.6$ & $33.6$ & $-$\\
	Yang et al. \cite{yang2014context} & $79.8$ & $48.7$ & $-$\\
	Shuai et al. \cite{shuai2016scene} & $81.2$ & $45.5$ & $-$\\\cline{1-4}
	ParseNet \cite{liu2017deep} & $86.8$ & $52.0$ & $40.4$\\
	ConvPP-8s \cite{xie2016top} & $-$ & $-$ & $40.7$\\
	FCN-8s \cite{shelhamer2017fully} & $85.9$ & $53.9$ & $41.2$\\
	DAG-RNN + CRF \cite{shuai2018scene} & $87.8$ & $57.8$ & $44.8$\\
	DAG-RNN \cite{shuai2018scene} & $87.3$ & $\mathbf{60.2}$ & $44.4$\\
	UoA-Context + CRF \cite{lin2016efficient} & $\mathbf{88.1}$ & $53.4$ & $\mathbf{44.9}$\\
	\bottomrule
      \end{tabularx}
    \begin{tablenotes}
	\item [\textbullet] Note: For fair comparison, all current methods below the middle horizontal line apply VGG-16 \cite{simonyan2014very} as visual perception module. While the previous methods still employ their default settings.
      \end{tablenotes}
    \end{threeparttable}
\end{table}

\begin{table}[!t]
   \centering
   \begin{threeparttable}
	\renewcommand{\arraystretch}{1.5}
	\caption{Performance comparison (\%) on COCO Stuff dataset (171 classes).}
	\label{vss_3}
	\centering
    \begin{tabularx}{\columnwidth}{bsss}
	\toprule
	{Method} & {GPA} & {ACA} & {mIOU}\\
	\midrule
	FCN \cite{caesar2016coco} & $52.0$ & $34.0$ & $22.7$\\
	DeepLab \cite{chen2015semantic} & $57.8$ & $38.1$ & $26.9$\\
	FCN-8s \cite{shelhamer2017fully} & $60.4$ & $38.5$ & $27.2$\\
	DAG-RNN \cite{shuai2018scene} & $62.2$ & $42.3$ & $30.4$\\
	DAG-RNN + CRF \cite{shuai2018scene} & $\mathbf{63.0}$ & $\mathbf{42.8}$ & $\mathbf{31.2}$\\
	\bottomrule
      \end{tabularx}
      \begin{tablenotes}
	\item [\textbullet] Note: For fair comparison, all the above methods apply VGG-16 \cite{simonyan2014very} as visual perception module.
      \end{tablenotes}
    \end{threeparttable}
\end{table}

Recently, due to the effective feature generation, CNN-based visual semantic segmentation methods are becoming popular. For instance, FCN \cite{caesar2016coco} and its variant FCN-8s \cite{shelhamer2017fully} are the most well-known examples. However, the direct prediction of those visual perception models generally are in low-resolution. To obtain high-resolution predictions, various visual semantic segmentation methods using visual context reasoning modules are proposed, i.e. DeepLab + CRF \cite{chen2015semantic}, CRF-RNN \cite{zheng2015conditional}, UoA-Context + CRF \cite{lin2016efficient} and DAG-RNN +CRF \cite{shuai2018scene}. Essentially, combing the strength of CNNs and CRFs for semantic segmentation becomes the focus. Among those methods, only DeepLab + CRF \cite{chen2015semantic} trains FCN \cite{caesar2016coco} and applies a dense CRF method as a post-processing step, while others jointly learn the dense CRFs and CNNs. Most of the above methods only incorporate pairwise (binary) potential terms within their corresponding variational free energies.

According to the comparison results shown in Table \ref{vss_1}, \ref{vss_2}, \ref{vss_3}, the recently proposed DAG-RNN + CRF \cite{shuai2018scene} method achieves the best performances in most scenarios while the previous UoA-Context + CRF \cite{lin2016efficient} algorithm only outperforms it in few cases within the \textit{Sift Flow} benchmark. This is because DAG-RNN module within the DAG-RNN + CRF \cite{shuai2018scene} method incorporates higher-order potential terms into the variational free energy instead of only applying pairwise potential terms like the previous methods. It is capable of enforcing local consistency and can enforce higher-order semantic coherence to a large extent \cite{shuai2018scene}. Moreover, the CRF module boosts the unary predictions and improves the ability of localizing object boundaries, which is inferior in DAG-RNN module \cite{shuai2018scene}. 
\begin{table}[!t]
   \resizebox{\columnwidth}{!}{
   \begin{threeparttable}
	\renewcommand{\arraystretch}{1.5}
	\caption{Performance comparison on Visual Relationship dataset ($k=1$).}
	\label{vrd_1}
	\centering
    \begin{tabular}{@{\extracolsep{4pt}}*7c@{}}
	\toprule
	{} & \multicolumn{2}{c}{Predicate} & \multicolumn{2}{c}{Phrase} & \multicolumn{2}{c}{Relationship}\\ \cmidrule{2-3} \cmidrule{4-5} \cmidrule{6-7}
	{Method} & {R@50} & {R@100} & {R@50} & {R@100} & {R@50} & {R@100}\\
	\midrule
	LP \cite{lu2016visual} & $47.87$ & $47.87$ & $16.17$ & $17.03$ & $13.86$ & $14.70$\\
	VTransE \cite{zhang2017visual} & $44.76$ & $44.76$ & $19.42$ & $22.42$ & $14.07$ & $15.20$\\
	PPRFCN \cite{zhang2017ppr} & $47.43$ & $47.43$ & $19.62$ & $23.15$ & $14.41$ & $15.72$\\
	SA-Full \cite{peyre2017weakly} & $50.40$ & $50.40$ & $16.70$ & $18.10$ & $14.90$ & $16.10$\\
	CAI \cite{zhuang2017towards} & $53.59$ & $53.59$ & $17.60$ & $19.24$ & $15.63$ & $17.39$\\
	ViP \cite{li2017vip} & $-$ & $-$ & $22.78$ & $27.91$ & $17.32$ & $20.01$\\
	VRL \cite{liang2017deep} & $-$ & $-$ & $21.37$ & $22.60$ & $18.19$ & $20.79$\\
	Zoom-Net \cite{yin2018zoom} & $50.69$ & $50.69$ & $24.82$ & $28.09$ & $18.92$ & $21.41$\\
	LK \cite{yu2017visual} & $55.16$ & $55.16$ & $23.14$ & $24.03$ & $19.17$ & $21.34$\\
	CAI + SCA-M \cite{yin2018zoom} & $\mathbf{55.98}$ & $\mathbf{55.98}$ & $\mathbf{25.21}$ & $\mathbf{28.89}$ & $\mathbf{19.54}$ & $\mathbf{22.39}$\\
	\bottomrule
    \end{tabular}
    \begin{tablenotes}
	\item [\textbullet] Note: All the above methods apply RPN \cite{ren2015faster} and triplet NMS \cite{li2017vip} to generate object proposals and remove redundant triplet candidates, respectively.
      \end{tablenotes}
    \end{threeparttable}
    }
\end{table} 

\subsection{Visual Relationship Detection}

In this section, two main benchmarks - \textit{visual relationship dataset} \cite{lu2016visual} and \textit{visual genome} \cite{krishna2017visual} - are used to compare different visual relationship detection methods on three tasks: \textit{predicate recognition} where both the bounding boxes and labels of the subject and object are given; \textit{phrase recognition} which predict the triple labels given a triplet structure as a union bounding box; \textit{relationship recognition} which also outputs triple labels but detects separate bounding boxes of the subject and object. Specifically, for \textit{visual relationship dataset}, 10 state-of-the-art visual relationship detection methods are chosen in the performance comparison including LP \cite{lu2016visual}, VTransE \cite{zhang2017visual}, CAI \cite{zhuang2017towards}, ViP \cite{li2017vip}, VRL \cite{liang2017deep}, LK \cite{yu2017visual}, PPRFCN \cite{zhang2017ppr}, SA-Full \cite{peyre2017weakly}, Zoom-Net \cite{yin2018zoom} and CAI + SCA-M \cite{yin2018zoom}, as shown in Table \ref{vrd_1}; For \textit{visual genome}, we compare three current visual relationship detection methods: DR-Net \cite{dai2017detecting}, ViP \cite{li2017vip} and Zoom-Net \cite{yin2018zoom}, as demonstrated in Table \ref{vrd_2}. Moreover, the evaluation metric $recall @ x$ \cite{alexe2012measuring} performance is relative to the number of predicates per subject-object pair to be evaluated, i.e. top $k$ predictions. In this survey, we choose $k=1$ for \textit{visual relationship dataset} and $k=100$ for \textit{visual genome}. The IOU between the predicated bounding boxes and the ground-truth is required above $0.5$ for the above methods. Furthermore, for a fair comparison, all methods mentioned above apply RPN \cite{ren2015faster} and triplet NMS \cite{li2017vip} to generate object proposals and remove redundant triplet candidates, respectively.
\begin{table}[!t]
   \resizebox{\columnwidth}{!}{
   \begin{threeparttable}
	\renewcommand{\arraystretch}{1.5}
	\caption{Performance comparison on Visual Genome dataset ($k=100$).}
	\label{vrd_2}
	\centering
    \begin{tabular}{@{\extracolsep{4pt}}*7c@{}}
	\toprule
	{} & \multicolumn{2}{c}{Predicate} & \multicolumn{2}{c}{Phrase} & \multicolumn{2}{c}{Relationship}\\ \cmidrule{2-3} \cmidrule{4-5} \cmidrule{6-7}
	{Method} & {R@50} & {R@100} & {R@50} & {R@100} & {R@50} & {R@100}\\
	\midrule
	DR-Net \cite{dai2017detecting} & $62.05$ & $71.96$ & $13.51$ & $17.23$ & $12.56$ & $16.06$\\
	ViP \cite{li2017vip} & $63.44$ & $74.15$ & $15.70$ & $19.96$ & $14.78$ & $18.85$\\
	Zoom-Net \cite{yin2018zoom} & $\mathbf{67.25}$ & $\mathbf{77.51}$ & $\mathbf{20.84}$ & $\mathbf{26.16}$ & $\mathbf{19.97}$ & $\mathbf{25.07}$\\
	\bottomrule
    \end{tabular}
    \begin{tablenotes}
	\item [\textbullet] Note: All the above methods apply RPN \cite{ren2015faster} and triplet NMS \cite{li2017vip} to generate object proposals and remove redundant triplet candidates, respectively.
      \end{tablenotes}
    \end{threeparttable}
    }
\end{table}  

According to the results shown in Table \ref{vrd_1}, the recently proposed CAI + SCA-M method \cite{yin2018zoom} outperforms previous visual relationship detection methods in all comparison criteria. For better understanding of the comparison results, we divide the methods into three different categories. Specifically, PPRFCN \cite{zhang2017ppr} and SA-Full \cite{peyre2017weakly} are essentially weakly-supervised visual relationship detection methods, that cannot generate comparable detection performance as other fully-supervised algorithms. Among all the fully-supervised methods, VTransE \cite{zhang2017visual}, ViP \cite{li2017vip} and Zoom-Net \cite{yin2018zoom} are virtually bottom-up visual relationship pursuit models, which only incorporate internal visual prior knowledge into the detecting procedure. Unlike the above algorithms, the top-down visual relationship pursuit methods such as LP \cite{lu2016visual}, CAI \cite{zhuang2017towards}, VRL \cite{liang2017deep}, LK \cite{yu2017visual} and CAI + SCA-M \cite{yin2018zoom} distill external linguistic prior knowledge into the learning frameworks. Generally, the external linguistic prior knowledge would regularize the original constraint optimization problems so that the associated top-down methods tend to bias towards certain feasible polytopes.

Unlike Table \ref{vrd_1}, all methods in Table \ref{vrd_2} are bottom-up visual relationship pursuit methods. We can observe in Table \ref{vrd_2} that Zoom-Net \cite{yin2018zoom} outperforms other two methods by a large margin, especially the visual relationship recognition task. As a visual semantic hierarchy reasoning model, Zoom-Net \cite{yin2018zoom} propagates contextual information among different visual semantic levels. Essentially, within the associated MAP inference, it can obtain a tighter upper bound for the target variational free energy and thus would generally converge to a better local optimum, as shown in Table \ref{vrd_2}.

\subsection{Scene Graph Generation}

In this section, seven available scene graph generation methods - IMP \cite{xu2017scene}, MSDN \cite{li2017scene}, NM-Freq \cite{zellers2018neural}, Graph R-CNN \cite{yang2018graph}, MotifNet \cite{zellers2018neural}, GPI \cite{herzig2018mapping} and LinkNet \cite{woo2018linknet}  - are compared on the \textit{visual genome} dataset \cite{krishna2017visual}, as shown in Table \ref{sgg}. Various visual genome dataset cleaning strategies exist in the current literature and, for a fair comparison, we choose the one used in the pioneering work \cite{xu2017scene} as the universal preprocessing model for all the above methods. Such cleaning strategy would generate training and test sets with 75,651 images and 32,422 images, respectively. Moreover, the most-frequent 150 object classes and 50 relation classes are selected in this survey. In general, each image has around 11.5 objects and 6.2 relationships in the scene graph. Furthermore, three evaluation aspects - \textit{Predicate Classification (PredCls)}, \textit{Phrase Classification (PhrCls)} and \textit{Scene Graph Generation (SGGen)} - are considered in this survey. Specifically, \textit{PredCls} represents the performance for recognizing the relation between two objects given the ground-truth locations; \textit{PhrCls} depicts the performance in the task of recognizing two object categories and their relation given the ground-truth locations; \textit{SGGen} indicates the performance for detecting objects (IOU $>$ 0.5) and recognising the predicates linking object pairs.
\begin{table}[!t]
   \resizebox{\columnwidth}{!}{
   \begin{threeparttable}
	\renewcommand{\arraystretch}{1.5}
	\caption{Performance comparison on Visual Genome dataset.}
	\label{sgg}
	\centering
    \begin{tabular}{@{\extracolsep{4pt}}*7c@{}}
	\toprule
	{} & \multicolumn{2}{c}{PredCls} & \multicolumn{2}{c}{PhrCls} & \multicolumn{2}{c}{SGGen}\\ \cmidrule{2-3} \cmidrule{4-5} \cmidrule{6-7}
	{Method} & {R@50} & {R@100} & {R@50} & {R@100} & {R@50} & {R@100}\\
	\midrule
	IMP \cite{xu2017scene} & $40.8$ & $45.2$ & $20.6$ & $22.4$ & $6.4$ & $8.0$\\
	MSDN \cite{li2017scene} & $53.2$ & $57.9$ & $27.6$ & $29.9$ & $7.0$ & $9.1$\\
	NM-Freq \cite{zellers2018neural} & $41.8$ & $48.8$ & $23.8$ & $27.2$ & $6.9$ & $9.1$\\
	Graph R-CNN \cite{yang2018graph} & $54.2$ & $59.1$ & $29.6$ & $31.6$ & $11.4$ & $13.7$\\
	MotifNet \cite{zellers2018neural} & $65.2$ & $67.1$ & $35.8$ & $36.5$ & $27.2$ & $\mathbf{30.3}$\\
	GPI \cite{herzig2018mapping} & $65.1$ & $66.9$ & $36.5$ & $38.8$ & $-$ & $-$\\
	LinkNet \cite{woo2018linknet} & $\mathbf{67.0}$ & $\mathbf{68.5}$ & $\mathbf{41.0}$ & $\mathbf{41.7}$ & $\mathbf{27.4}$ & $30.1$\\
	\bottomrule
    \end{tabular}
    \begin{tablenotes}
	\item [\textbullet] Note: All the above methods apply the same cleaning strategy proposed in paper \cite{xu2017scene}.
      \end{tablenotes}
    \end{threeparttable}
    }
\end{table} 

Unlike visual relationship detection, scene graph generation needs to model global inter-dependency among the entire object instances, rather than focus on local relationship triplets in isolation. Essentially, the strong independence assumptions in local predictors limit the quality of the global predictions \cite{zellers2018neural}. As shown in Table \ref{sgg}, the first four methods (IMP \cite{xu2017scene}, MSDN \cite{li2017scene}, NM-Freq \cite{zellers2018neural} and Graph R-CNN \cite{yang2018graph}) use graph-based inference to propagate local contextual information in both directions between object and relationship nodes, while the last three methods (MotifNet \cite{zellers2018neural}, GPI \cite{herzig2018mapping} and LinkNet \cite{woo2018linknet}) tend to incorporate global contextual information within the inference procedure. From Table \ref{sgg}, it can be seen that the latter methods incorporating global contextual information outperform the previous ones to a large extent. Among them, the recently proposed LinkNet \cite{woo2018linknet} achieves the best performance in almost all comparison criteria. This is mainly because the authors propose a simple and effective relational embedding module to explicitly model the global contextual information. 

\section{Conclusion}

This survey presents a comprehensive review of state-of-the-art visual semantic information pursuit methods. Specifically, we mainly focus on four related applications: object detection, visual semantic segmentation, visual relationship detection and scene graph generation. To understand the essence of the current visual semantic information pursuit methods, a specific unified paradigm is distilled in this survey. The main developments and the future trends in each potential direction are also reviewed, followed by summarising the most popular benchmarks, the evaluation metrics, and the relative performance of the key algorithms. 

% or


%\appendices

%\section{}

\ifCLASSOPTIONcompsoc
  \section*{Acknowledgments}
\else
  \section*{Acknowledgment}
\fi

This work was supported in part by the U.K. Defence Science and Technology Laboratory, and in part by the Engineering and Physical Research Council (collaboration between U.S. DOD, U.K. MOD, and U.K. EPSRC through the Multidisciplinary University Research Initiative) under Grant EP/K014307/1 and Grant EP/R018456/1.

\ifCLASSOPTIONcaptionsoff
  \newpage
\fi

\bibliographystyle{IEEEtran}
\bibliography{IEEEabrv,Semantic}

%\begin{thebibliography}{1}


%\end{thebibliography}


%\end{IEEEbiography}


%\newpage



%\vfill


\end{document}